\documentclass[11pt,a4paper]{article}
\usepackage[hyperref]{acl2020}
\usepackage{times}
\usepackage{graphicx}
\usepackage{latexsym}
\usepackage{booktabs}
\usepackage{amssymb}
\usepackage{amsmath}
\usepackage{lipsum}

\usepackage{pifont}
\usepackage{url}
\usepackage{bm}
\usepackage{multirow}
\usepackage{pstool}
\DeclareMathOperator*{\Softmax}{Softmax}

\DeclareMathOperator*{\Transformer}{Transformer}
\DeclareMathOperator*{\LayerNorm}{LayerNorm}

\DeclareMathOperator*{\Tanh}{Tanh}
\DeclareMathOperator*{\GELU}{GELU}
\DeclareMathOperator*{\Sigmoid}{Sigmoid}
\DeclareMathOperator*{\MeanPooling}{MeanPooling}
\DeclareMathOperator*{\MaxPooling}{MaxPooling}

\DeclareMathOperator*{\argmaxK}{argmax-K}

\newcommand{\real}[1]{\mathbb{R}^{#1}}
\newcommand{\transpose}[1]{#1^{\top}}
\usepackage{microtype}
\usepackage{colortbl}
\definecolor{gray}{RGB}{87, 87, 87}
\definecolor{red}{RGB}{173, 35, 35}
\definecolor{blue}{RGB}{42, 75, 215}
\definecolor{green}{RGB}{29, 105, 20}
\definecolor{brown}{RGB}{129, 74, 25}
\definecolor{purple}{RGB}{129, 38, 192}
\definecolor{cyan}{RGB}{41, 208, 208}
\definecolor{yellow}{RGB}{189, 167, 0}
\definecolor{Red}{rgb}{0.68, 0.05, 0.0}
\definecolor{Blue}{rgb}{0.0, 0.0, 0.61}
\definecolor{Blue1}{RGB}{214, 235, 245}
\definecolor{Blue2}{RGB}{235, 245, 250}
\definecolor{lime}{RGB}{60,179,113}
\definecolor{peach}{RGB}{255, 242, 230}
\aclfinalcopy 
\def\aclpaperid{1043} 

\title{RikiNet: Reading Wikipedia Pages for Natural Question Answering}
\author{
Dayiheng Liu\footnotemark[2]\hspace{1mm}\thanks{\hspace{2mm}Work is done during internship at Microsoft Research Asia.} , Yeyun Gong\footnotemark[3], Jie Fu\footnotemark[4], Yu Yan\footnotemark[5], Jiusheng Chen\footnotemark[5], Daxin Jiang\footnotemark[6], Jiancheng Lv\footnotemark[2], Nan Duan\footnotemark[3] \\
\footnotemark[2]\hspace{0.5mm} College of Computer Science, Sichuan University\\
\footnotemark[3]\hspace{0.5mm} Microsoft Research Asia \\
\footnotemark[4]\hspace{0.5mm} Mila, Polytechnique Montreal\\
\footnotemark[5]\hspace{0.5mm} Microsoft AI and Research \\
\footnotemark[6]\hspace{0.5mm} Microsoft Search Technology Center Asia\\
{\tt losinuris@gmail.com
 }
 } 

\date{}

\begin{document}
\maketitle
\begin{abstract}
Reading long documents to answer open-domain questions remains challenging in natural language understanding. 
In this paper, we introduce a new model, called RikiNet, which reads Wikipedia pages for natural question answering. 
RikiNet contains a dynamic paragraph dual-attention reader and a multi-level cascaded answer predictor. 
The reader dynamically represents the document and question by utilizing a set of complementary attention mechanisms. 
The representations are then fed into the predictor to obtain the span of the short answer, the paragraph of the long answer, and the answer type in a cascaded manner. 
On the Natural Questions (NQ) dataset, a single RikiNet achieves 74.3 F1 and 57.9 F1 on long-answer and short-answer tasks. To our best knowledge, it is the first single model that outperforms the single human performance.
Furthermore, an ensemble RikiNet obtains 76.1 F1 and 61.3 F1 on long-answer and short-answer tasks, achieving the best performance on the official NQ leaderboard\footnote{Till our submission time, 29 Nov. 2019. We refer readers to \url{https://ai.google.com/research/NaturalQuestions/leaderboard} for the latest results.}.  
\end{abstract}


\section{Introduction}
Machine reading comprehension (MRC) refers to the task of finding answers to given questions by reading and understanding some documents. 
It represents a challenging benchmark task in natural language understanding (NLU). 
With the progress of large-scale pre-trained language models~\cite{devlin2018bert}, state-of-the-art MRC models~\cite{ju2019technical,yang2019xlnet,lan2019albert,zhang2019sg,liu2019roberta} have already surpassed human-level performance on certain commonly used MRC benchmark datasets, such as SQuAD 1.1~\cite{rajpurkar2016squad}, SQuAD 2.0~\cite{rajpurkar2018know}, and CoQA~\cite{reddy2019coqa}.

Recently, a new benchmark MRC dataset called Natural Questions\footnote{NQ provides some visual examples of the data at \url{https://ai.google.com/research/NaturalQuestions/visualization}.} (NQ)~\cite{kwiatkowski2019natural} has presented a substantially greater challenge for the existing MRC models.
Specifically, there are two main challenges in NQ compared to the previous MRC datasets like SQuAD 2.0. 
\textbf{Firstly}, instead of providing one relatively short paragraph for each question-answer (QA) pair, NQ gives an entire Wikipedia page which is significantly longer compared to other datasets.
\textbf{Secondly}, NQ task not only requires the model to find an answer span (called short answer) to the question like previous MRC tasks but also asks the model to find a paragraph that contains the information required to answer the question (called long answer).

In this paper, we focus on the NQ task and propose a new MRC model called \textbf{RikiNet} tailored to its associated challenges, which \textbf{R}eads the W\textbf{iki}pedia pages for natural question answering. 
For the first challenge of the NQ task mentioned above, RikiNet employs the proposed Dynamic Paragraph Dual-Attention (DPDA) reader which contains multiple DPDA blocks. 
In each DPDA block, we iteratively perform dual-attention to represent documents and questions, and employ paragraph self-attention with dynamic attention mask to fuse key tokens in each paragraph. 
The resulting context-aware question representation, question-aware token-level, and paragraph-level representations are fed into the predictor to obtain the answer.
The motivations of designing DPDA reader are: (a) Although the entire Wikipedia page contains a large amount of text, one key observation is that most answers are only related to a few words in one paragraph; (b) The final paragraph representation can be used naturally for predicting long answers. 
We describe the details of DPDA reader in \S~\ref{sec:dpcr}.

For the second challenge, unlike prior works on NQ dataset~\cite{alberti2019bert,pan2019frustratingly} that only predict the short answer and directly select its paragraph as long answer, RikiNet employs a multi-level cascaded answer predictor which jointly predict the short answer span, the long answer paragraph, and the answer type in a cascaded manner.
Another key intuition motivating our design is that even if the relevant documents are not given, humans can easily judge that some questions have no short answers \cite{benjamin2019}. 
Take this question as a motivating example:``What is the origin of the Nobel prize?'' 
The answer should be based on a long story, which cannot be easily expressed in a short span of entities.  
Therefore we also feed the question representation into the predictor as an auxiliary prior to answer type prediction.
The details will be given in \S~\ref{sec:mcap}.

On the NQ test set, our single model obtains 74.3 F1 scores on the long-answer task (LA) and 57.9 F1 scores on the short-answer task (SA) compared to the published best single model~\cite{alberti2019synthetic} results of 66.8 F1 on LA and 53.9 F1 on SA. 
To the best of our knowledge, RikiNet is the first \textit{single} model that outperforms the single human performance~\cite{kwiatkowski2019natural} on both LA and SA. 
Besides, our ensemble model obtains 76.1 F1 on LA and 61.3 F1 on SA, which achieves the best performance of both LA and SA on the official NQ leaderboard. 

\section{Preliminaries} \label{sec:3.1}
Before we describe our model in detail, we first introduce the notations and problem formalization.
Our paper considers the following NQ~\cite{kwiatkowski2019natural} task:
Given a natural question $q$, a related Wikipedia page $p$ (in the top 5 search results returned by the Google search engine), the model outputs a paragraph within the Wikipedia page $p$ as the \textit{long answer} which contains enough information to infer the answer to the question, and an entity span within the long answer that answers the question as the \textit{short answer}. Also, the short answer of the 1\% Wikipedia page is ``yes'' or ``no'', instead of a short span. Both long answers and short answers can be $\mathtt{NULL}$ (\textit{i.e.}, no such answer could be found).

Given a natural question $q$ and its paired Wikipedia page $p$, we tokenize them with the 30,522 wordpiece vocabulary as used in~\cite{devlin2018bert}. Following~\cite{alberti2019bert,pan2019frustratingly}, we generate multiple document spans by splitting the Wikipedia page with a sliding window. Then, we obtain multiple 6-tuple training instances $(q,d,c,s,e,t)$ for each NQ data pair $(q,p)$, where $q$ and $d$ are wordpiece IDs of question with length $n$ and document span with length $m$, $c\in \mathbb{S}$ indicates the paragraph index of the long answer where $\mathbb{S}$ is the set that includes all paragraph indexes (\textit{i.e,} all long answer candidates) within $d$, $s,e \in \{0,1,...,m-1\}$ are inclusive indices pointing to the start and end of the short answer span, and $t \in \{0,1,2,3,4\}$ represents the five answer types, corresponding to the labels ``NULL'' (no answer), ``SHORT'' (has short answer), ``LONG'' (only has long answer), ``YES'', and ``NO''.

For each tuple $(q,d,c,s,e,t)$ of the data pair $(q,p)$, RikiNet takes $d$ and $q$ as inputs, and jointly predicts $c,s,e,t$. Finally we merge the prediction results of every tuple to obtain the final predicted long answer, short answer, and their confidence scores of the data pair $(q,p)$ for evaluation.

\begin{figure*}[t] 
  \centering
  \includegraphics[scale=0.49]{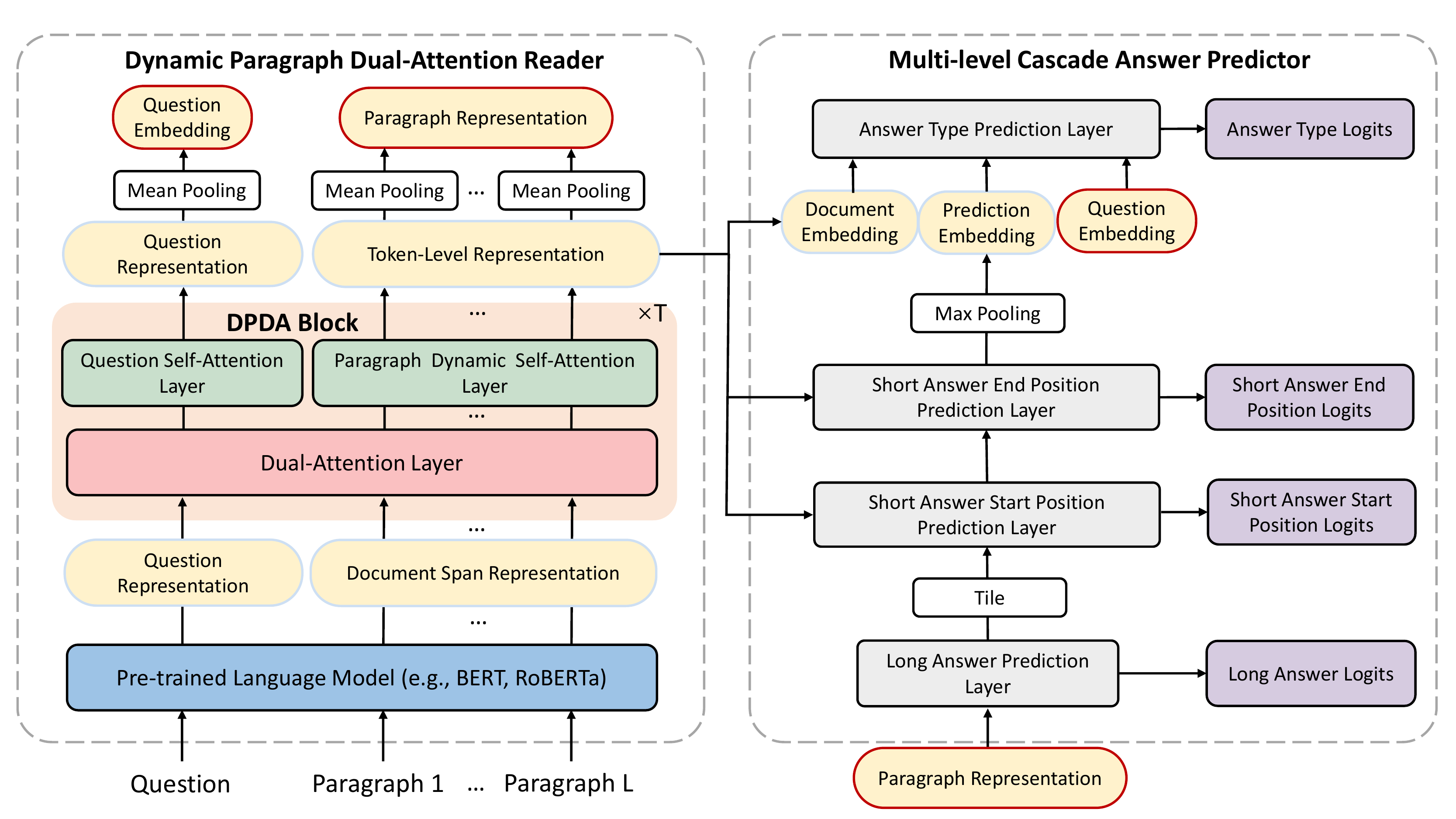}
  \caption{Overview of RikiNet framework.} \label{fig:model}
\end{figure*}

\section{Methodology}
We propose the \textbf{RikiNet} which \textbf{R}eads the W\textbf{iki}pedia pages for natural question answering. As shown in Fig.~\ref{fig:model}, RikiNet consists of two modules: (a) the \textbf{dynamic paragraph dual-attention reader} as described in \S\ref{sec:dpcr}, and (b) the \textbf{multi-level cascaded answer predictor} as described in \S\ref{sec:mcap}. 

\subsection{Dynamic Paragraph Dual-Attention Reader} \label{sec:dpcr}
Dynamic Paragraph Dual-Attention (DPDA) reader aims to represent the document span $d$ and the question $q$. 
It outputs the context-aware question representation, question-aware token-level document representation, and paragraph-level document representation, which will be all fed into the predictor to obtain the long and short answers.

\subsubsection{Encoding Question and Document Span} \label{sec:3.2.1}
We firstly employ a pre-trained language model such as BERT~\cite{devlin2018bert} to obtain the initial question representation $Q_0 \in \real{n \times h}$ and the initial document span representation $D_0 \in \real{m \times h}$, where $h$ is the hidden size.
Similar to~\cite{devlin2018bert}, we  concatenate a ``[CLS]'' token, the tokenized question $q$ with length $n$, a  ``[SEP]'' token, the tokenized document span $d$ with length $m$, and a final ``[SEP]'' token. Then we feed the resulting sequence into the pre-trained language model.

\subsubsection{Dynamic Paragraph Dual-Attention Block}
As shown on the left in Fig.~\ref{fig:model}, DPDA reader contains multiple Dynamic Paragraph Dual-Attention (DPDA) blocks. The first block takes $Q_0$ and $D_0$ as the inputs. The outputs $Q_{(t)}$ and $D_{(t)}$ of the $t$-th block are then fed into the next block. Each block contains three types of layers: the dual-attention layer, the paragraph dynamic self-attention layer, and the question self-attention layer. The last DPDA block outputs the final question and document representations. 
We describe them in detail now. 

\paragraph{Dual-Attention Layer}
To strengthen the information fusion from the question to the paragraphs as well as from the paragraphs to the question, we adapt a dual-attention mechanism, which has been shown effective in other MRC models~\cite{xiong2017dcn+,seo2016bidirectional,xiong2016dynamic}.
We further tweak it by increasing the depth of attention followed by a residual connection~\cite{he2016deep} and layer normalization~\cite{ba2016layer}. 

In particular, the $t$-th block first calculates a similarity metric $L_{(t)} \in \real{m \times n}$ which is then normalized row-wise and column-wise to produce two attention weights: $A^Q_{(t)} \in \real{m \times n}$, across the document for each token in the question; and $A^D_{(t)} \in \real{n \times m}$, across the question for each token in the document,
\begin{align*}
    L_{(t)} &= D_{(t-1)} \transpose{Q}_{(t-1)} \in \real{m \times n}, \\
    A^Q_{(t)} &= \Softmax \left( L_{(t)} \right) \in \real{m \times n}, \\
    A^D_{(t)} &= \Softmax \left(\transpose{L}_{(t)} \right) \in \real{n \times m}. 
\end{align*}
Similar to ~\cite{xiong2016dynamic,seo2016bidirectional}, we obtain the question-aware representation of the document by
\begin{align*}
    \bar{\bar{Q}}_{(t)}^C &= \transpose{\left(\transpose{D}_{(t-1)} A_{(t)}^Q\right)}   \in \real{n \times h}, \\
    \bar{D}_{(t)}^C &= \transpose{\left(A^D_{(t)}\right)} \left[ Q_{(t-1)} ; \bar{\bar{Q}}_{(t)}^C \right]  \in \real{m \times 2 h},
\end{align*}
where $[\cdot;\cdot]$ denotes concatenation. 
We also obtain the context-aware question representation in a dual way:
\begin{align*}
    \bar{\bar{D}}_{(t)}^C &= \transpose{\left(\transpose{Q_{(t-1)}} A^D_{(t)}\right)}   \in \real{m \times h}, \\
    \bar{Q}_{(t)}^C &= \transpose{\left(A_{(t)}^Q \right)} \left[ D_{(t-1)} ; \bar{\bar{D}}_{(t)}^C \right]  \in \real{n \times 2 h}.
\end{align*}
We finally apply the residual connection and layer normalization to both the question and the document representations with the linear transformations.
\begin{align*}
      D^C_{(t)} &= \LayerNorm\left(D_{(t-1)} + \bar{D}_{(t)}^CW_{(t)}^D\right) \in \real{m \times h}, \\
      Q^C_{(t)} &= \LayerNorm\left(Q_{(t-1)} + \bar{Q}_{(t)}^CW_{(t)}^Q\right) \in \real{n \times h}, 
\end{align*}
where $W^D_{(t)} \in \real{2 h \times h}$ and $W^Q_{(t)} \in \real{2 h \times h}$ are trainable parameters in the dual-attention layer of the $t$-th block. 
The document representation $D^C_{(t)}$ will be fed into the paragraph dynamic self-attention layer to obtain the paragraph representation.
The question representation $Q^C_{(t)}$ will be fed into the question self-attention layer to get the question embedding. 

\paragraph{Question Self-Attention Layer}
This layer uses a transformer self-attention block~\cite{vaswani2017attention} to further enrich the question representation:  
\begin{align*}
     Q_{(t)} &= \Transformer\left(Q^C_{(t)}\right) \in \real{n \times h},
\end{align*}
where the transformer block consists of two sub-layers: a multi-head self-attention layer and a position-wise fully connected feed-forward layer. Each sub-layer is placed inside a residual connection with layer normalization.
After the last DPDA block, we obtain the final question embedding $\bm{q} \in \real{h}$ by applying the mean pooling, 
\begin{align*}
     \bm{q} &= \MeanPooling\left(Q^C_{(T)}\right) \in \real{h},
\end{align*}
where $T$ denotes the number of the DPDA blocks. This question embedding $\bm{q}$ will be further fed into the predictor for answer type prediction.

\paragraph{Paragraph Dynamic Self-Attention Layer}
This layer is responsible for gathering information on the key tokens in each paragraph. 
The token-level representation $D_{(t)}$ is first given by:

\begin{align}
     D_{(t)} &= \Transformer\left(D^C_{(t)}\right) \in \real{m \times h}. \label{eq:1}
\end{align}
The difference from the original multi-head self-attention in~\cite{vaswani2017attention} is that we incorporate two extra attention masks, which will be introduced later in Eq. (\ref{eq:lmask}) and (\ref{eq:dmask}).
The last DPDA block applies a mean pooling to the tokens within the same paragraph to obtain the paragraph representation $L \in \real{l \times h}$ as
\begin{align}
     L[i,:] &= \MeanPooling\limits_{\mathbb{L}_j= i}\left(\left\{D_{(T)}[j,:]\right\}\right) \in \real{h}, \label{eq:2}
\end{align}
where $l$ denotes the number of paragraph within the document span $d$ (\textit{i.e.}, the number of long answer candidates within the document span $d$), $L[i,:]$ is the representation of the $i$-th paragraph, $D_{(T)}[j,:]$ is the representation of the $j$-th token at last DPDA block, and ${\mathbb{L}_j}$ indicates the index number of the paragraph where the $j$-th token is located. 

Tokens in the original multi-head attention layer of the transformer self-attention block attend to all tokens.  
We introduce two attention masks to the self-attention sub-layer in Eq. (\ref{eq:1}) based on two key motivations: 1) Each paragraph representation should focus on the question-aware token information inside the paragraph; 2) Most of the answers are only related to a few words in a paragraph.
For the first motivation, we introduce the paragraph attention mask $\mathcal{M}^L \in \real{m \times m}$ which is defined as:
\begin{align} \label{eq:lmask}
    \mathcal{M}^L[i,j] = \left\{\begin{matrix}
0, & {\text{ if } \mathbb{L}_i} = {\mathbb{L}_j},  \\ 
-\infty, & {\text{otherwise}}.
\end{matrix}\right.
\end{align}
It forces each token to only attend to the tokens within the same paragraph. Therefore, each paragraph representation focuses on its internal token information after the mean pooling of Eq. (\ref{eq:2}).

Based on the second motivation, we dynamically generate another attention mask to select key tokens before self-attention. We use a neural network $\mathcal{F}_{(t)}^\Phi$ called scorer with the $\Sigmoid$ activation function to calculate the importance score for each token:
\begin{align*}
    \Phi_{(t)} = \mathcal{F}_{(t)}^\Phi\left(D^C_{(t)}\right) \in \real{m \times 1},
\end{align*}
Then we obtain the dynamic attention mask $\mathcal{M}_{(t)}^\Phi \in \real{m \times m}$ by selecting top-$K$ tokens\footnote{Following~\citet{zhuang2019token}, our implementation pads the unselected token representations with zero embeddings and adds the scorer representation with the linear transformation to $D_{(t)}$ to avoid gradient vanishing for scorer training.}
\begin{align} \label{eq:dmask}
    \mathcal{M}_{(t)}^\Phi[i,j] = \left\{\begin{matrix}
0, & {\text{ if } i \in {\mathbb{S}^\Phi_{(t)}} \text{ and } j \in {\mathbb{S}^\Phi_{(t)}}} \\ 
-\infty, & {\text{otherwise}},
\end{matrix}\right.
\end{align}
where $\mathbb{S}^\Phi_{(t)} = \argmaxK\limits_{k \in [0, m-1]} \left(\left\{\Phi_{(t)}[k]\right\}\right)$. 
Here $\Phi_{(t)}[k]$ denotes the score of the $k$-th token at $t$-th block, $K$ is a hyperparameter, and $\mathbb{S}^\Phi_{(t)}$ is the set that includes the index of the selected top-$K$ tokens. This attention mask lets the paragraph representation concentrate on the selected key tokens.

The final scaled dot-product attention weight $A_{(t)} \in \real{m \times m}$ of the multi-head self-attention sub-layer~\cite{vaswani2017attention} in Eq. (\ref{eq:1}) with two proposed attention masks can be written as:
\begin{align*} 
A_{(t)} = \Softmax \left(\mathcal{M}_{(t)}^\Phi+\mathcal{M}^L+  {\frac{{\left( {D^C_{(t)}}\transpose{{D^C_{(t)}}} \right)}}{{\sqrt {{h}} }}} \right).
\end{align*}

\subsection{Multi-level Cascaded Answer Predictor}\label{sec:mcap}
Due to the nature of the NQ tasks, a short answer is always contained within a long answer, and thus it makes sense to use the prediction of long answers to facilitate the process of obtaining short answers. 
As shown on the right in Fig.~\ref{fig:model}, we design a cascaded structure to exploit this dependency. 
This predictor takes the token representation $D_{(T)}$, the paragraph representation $L$, and the question embedding $\bm{q}$ as inputs to predict four outputs in a \textit{cascaded} manner: (1) long answer $\rightarrow$ (2) the start position of the short answer span $\rightarrow$ (3) the end position of the short answer span $\rightarrow$ (4) the answer type.
That is, the previous results are used for the next tasks as indicated by the notation ``$\rightarrow$''.

\paragraph{Long Answer Prediction} We employ a dense layer $\mathcal{F}^L$ with $\Tanh$ activation function as long answer prediction layer, which takes the paragraph representation $L \in \real{l \times h}$ as input to obtain the long-answer prediction representation $H^L \in \real{l \times h}$. Then the long-answer logits $\bm{o}^L$ are computed with a linear layer
\begin{align*} 
   H^L &= \mathcal{F}^L\left(L\right) \in \real{l \times h}, \\
   \bm{o}^L &= H^L W^L \in \real{l},
\end{align*}
where $W^L \in \real{h \times 1}$ is a trainable parameter.

\paragraph{Short Answer Prediction}
 
Firstly, we use the long-answer prediction representation $H^L$ and the token representation $D_{(T)}$ as the inputs to predict the start position of the short answer. Then the prediction representation of the start position of the short answer will be \textit{re-used} to predict the end position. 

Since the row-dimension of $D_{(T)} \in \real{m \times h}$ is different from that of $H^L \in \real{l \times h}$, we cannot directly concatenate the $H^L$ to $D_{(T)}$. 
We tile the $H^L \in \real{l \times h}$ with $\bar{H}^L \in \real{m \times h}$ along the row-dimension: $\bar{H}^L\left[i,:\right] = H^L\left[\mathbb{L}_i,:\right] \in \real{h}$. 
Note that ${\mathbb{L}_i}$ indicates the index number of the paragraph where the $i$-th token is located. 
Thus, the model can consider the prediction information of the long answer when predicting the short answer.
Similarly, the start and end position logits of the short answer are predicted by,
\begin{align*}
    &H^S = \mathcal{F}^S\left(\left[\bar{H}^L;D_{(T)}\right]\right) \in \real{m \times h}, \\
    &\bm{o}^S = H^S W^S \in \real{m}, \\
    &H^E = \mathcal{F}^E\left(\left[H^S;D_{(T)}\right]\right) \in \real{m \times h}, \\
    &\bm{o}^E = H^E W^E \in \real{m},
\end{align*}
where $\bm{o}^S$ and $\bm{o}^E$ are the output logit vectors of the start positions and the end positions of the short answer, $\mathcal{F}^S$ and $\mathcal{F}^E$ are two dense layers with $\Tanh$ activation function, and $W^S \in \real{h \times 1}$, $W^E \in \real{h \times 1}$ are trainable parameters.

\paragraph{Answer Type Prediction}
Finally, the predictor outputs the answer type. There are five answer types as discussed in \S~\ref{sec:3.1}. With the observation that humans can easily judge that some questions have no short answers even without seeing the document, we treat the question embedding $\bm{q} \in \real{h}$ as an auxiliary input for the answer type prediction. Besides, the token representation $D_{(T)}$ and the short-answer prediction representation $H^E$ are also used for that prediction:
\begin{align*}
    &\bm{d} = \MeanPooling\left(D_{(T)}\right) \in \real{h}, \\
    &\bm{e} = \MaxPooling\left(H^E\right) \in \real{h}, \\
    &\bm{h}^T = \mathcal{F}^T\left(\left[\bm{d};\bm{q};\bm{e}\right]\right) \in \real{h}, \\
    &\bm{o}^T = \Softmax\left(\bm{h}^T W^T\right) \in \real{5},
\end{align*}
where $\bm{o}^T$ is the logits of the five answer types, $\mathcal{F}^T$ is a dense layer with $\Tanh$ activation function, and $W^T \in \real{h \times 5}$ is a trainable parameter.

\paragraph{Training Loss and Inference}
For training, we compute cross-entropy loss over the above mentioned output logits, and jointly minimize these four cross-entropy losses as:
\begin{align*}
    \mathcal{L} = \mathcal{L}^L + \mathcal{L}^S+ \mathcal{L}^E + \mathcal{L}^T.
\end{align*}
During inference, we calculate the final long-answer score $\Psi^L$ for all the paragraphs within the Wikipedia page based on the long-answer logits $\bm{o}^L$ and the answer type logits $\bm{o}^T$. The long-answer score of paragraph $c$ can be written as
\begin{align*}
    \Psi^L(c) = \bm{o}^L[c] + \underbrace{\left(\sum_{t=1}^{4}\bm{o}^T[t]- \bm{o}^T[0]\right)}_{\text{answer type score}}, 
\end{align*}
where $\bm{o}^T[0]$ denotes the logits where the answer type is ``NULL''(no answer), $\sum_{t=1}^{4}\bm{o}^T[t]$ denotes the sum of the logits where the answer type is not ``NULL''. The answer type score can be seen as a bias of each document span in the Wikipedia page. Then we select the paragraph of the highest long-answer score $\Psi^L$ over the entire Wikipedia page as the long answer. 

Similarly, the short-answer score of the corresponding span $(s,e)$ is calculate by
\begin{align*}
    \Psi^S(s,e) = \underbrace{\left(\bm{o}^S[s] + \bm{o}^E[e]\right)}_{\text{answer span score}} + \underbrace{\left(\bm{o}^T\left[1\right] - \bm{o}^T[0]\right)}_{\text{answer type score}}, 
\end{align*}
where $\bm{o}^T[1]$ denotes the score where the answer type is ``SHORT''(has short answer). We select the short answer span which has the highest short-answer score $\Psi^S$ within the long answer as the final short answer. We use the official NQ evaluation script to set two separate thresholds for predicting whether the two types of answers are answerable. 

\section{Experiments}
\begin{table*}[ht]
\begin{center}
\resizebox{\textwidth}{!}{%
 \begin{tabular}{l c c c c c c c c c c c c c} 
 \toprule
   & \multicolumn{3}{c}{LA Dev}
   & \multicolumn{3}{c}{LA Test} &
   & \multicolumn{3}{c}{SA Dev}
   & \multicolumn{3}{c}{SA Test} \\
   &    P &    R &   F1 &    P &    R &   F1 &
   &    P &    R &   F1 &    P &    R &   F1 \\
 \midrule
   DocumentQA~\cite{clark2018simple} 
   & 47.5 & 44.7 & \cellcolor{Blue2}{46.1} & 48.9 & 43.3 & \cellcolor{Blue2}{45.7} &
   & 38.6 & 33.2 & \cellcolor{Blue2}{35.7} & 40.6 & 31.0 & \cellcolor{Blue2}{35.1} \\
   DecAtt~\cite{parikh2016decomposable} + DocReader~\cite{chen2017reading} 
   & 52.7 & 57.0 & \cellcolor{Blue2}{54.8} & 54.3 & 55.7 & \cellcolor{Blue2}{55.0} &
   & 34.3 & 28.9 & \cellcolor{Blue2}{31.4} & 31.9 & 31.1 & \cellcolor{Blue2}{31.5} \\
   BERT\textsubscript{joint}~\cite{alberti2019bert}
   & 61.3 & 68.4 & \cellcolor{Blue2}{64.7} & 64.1 & 68.3 & \cellcolor{Blue2}{66.2} &
   & 59.5 & 47.3 & \cellcolor{Blue2}{52.7} & 63.8 & 44.0 & \cellcolor{Blue2}{52.1} \\
   BERT\textsubscript{large} + 4M synth NQ ~\cite{alberti2019synthetic} 
   & 62.3 & 70.0 & \cellcolor{Blue2}{65.9} & 65.2  & 68.4 & \cellcolor{Blue2}{66.8} & 
   & 60.7 & 50.4 & \cellcolor{Blue2}{55.1} & 62.1 & 47.7 & \cellcolor{Blue2}{53.9}  \\
   BERT\textsubscript{joint}~\cite{alberti2019bert} + RoBERTa\textsubscript{ large}~\cite{liu2019roberta} $\ddag$
   & 65.6 & 69.1 & \cellcolor{Blue2}{67.3} & - & -  & - &
   & 60.9 & 51.0 & \cellcolor{Blue2}{55.5} & - & -  & -  \\
   BERT\textsubscript{large} + SQuAD2 PT + AoA~\cite{pan2019frustratingly}$\dagger$ 
   & - & - & \cellcolor{Blue2}{68.2} & - & -  & - &
   & - & - & \cellcolor{Blue2}{57.2} & -& -  & -  \\
   BERT\textsubscript{large} + SSPT~\cite{glass2019span}$\dagger$ 
   & - & - & \cellcolor{Blue2}{65.8} & - & -  & - &
   & - & - & \cellcolor{Blue2}{54.2} & - & -  & -  \\
   RikiNet-BERT\textsubscript{large} 
   & 73.2 & 74.5 & \cellcolor{Blue2}{73.9} & 74.2 & 74.4 & \cellcolor{Blue2}{74.3} &
   & 61.1 & 54.7 & \cellcolor{Blue2}{57.7} & 63.5 & 53.2 & \cellcolor{Blue2}{57.9}  \\
   RikiNet-RoBERTa\textsubscript{ large} $\ddag$
   & 74.3 & 76.4 & \cellcolor{Blue2}{\textbf{75.3}} & - & -  & - &
   & 61.4 & 57.3 & \cellcolor{Blue2}{\textbf{59.3}} & -& -  & -  \\
   \midrule
   RikiNet-BERT\textsubscript{large} (ensemble)
   & 74.4 & 76.3 & \cellcolor{Blue2}{75.4} & 75.3 & 75.9 & \cellcolor{Blue2}{75.6} &
   & 66.9 & 53.8 & \cellcolor{Blue2}{59.6} & 63.2 & 56.1 & \cellcolor{Blue2}{59.5}  \\
   RikiNet-RoBERTa\textsubscript{ large} (ensemble)
   & 73.3 & 78.7 & \cellcolor{Blue2}{\textbf{75.9}} & 78.1 & 74.2 & \cellcolor{Blue2}{\textbf{76.1}} &
   & 66.6 & 56.4 & \cellcolor{Blue2}{\textbf{61.1}} & 67.6 & 56.1  & \cellcolor{Blue2}{\textbf{61.3}}  \\
 \midrule
   Single Human~\cite{kwiatkowski2019natural}
   & 80.4 & 67.6 & \cellcolor{Blue2}{73.4} & -    & -    & -    &
   & 63.4 & 52.6 & \cellcolor{Blue2}{57.5} & -    & -    & -    \\
   Super-annotator~\cite{kwiatkowski2019natural}
   & 90.0 & 84.6 & \cellcolor{Blue2}{87.2} & -    & -    & -    &
   & 79.1 & 72.6 & \cellcolor{Blue2}{75.7} & -    & -    & -    \\
 \bottomrule
\end{tabular}}
\end{center}
\caption{Performance comparisons on the dev set and the blind test set of the NQ dataset. We report the evaluation results of the precision (P), the recall (R), and the F1 score for both long-answer (LA) and short-answer (SA) tasks. We use background color to highlight the column of F1 results. $\dagger$ refers to the works that only provide the F1 results on the dev set in their paper. $\ddag$ refers to our implementations where we only report the results on the dev set, due to the NQ leaderboard submission rules (each participant is only allowed to submit once per week).}
\label{tab:main}
\end{table*}

\subsection{Dataset}
We focus on the Natural Questions (NQ)~\cite{kwiatkowski2019natural} dataset in this work. The public release of the NQ dataset consists of 307,373 training examples and 7,830 examples for development data (dev set). NQ provides a blind test set contains 7,842 examples, which can only be accessed through a public leaderboard submission. 

\subsection{Implementation Details}
As discussed in \S~\ref{sec:3.1}, we generate multiple document spans by splitting the Wikipedia page with a sliding window. Following~\cite{pan2019frustratingly,alberti2019bert}, the size and stride of the sliding window are set to 512 and 192 tokens respectively. The average number of document spans of one Wikipedia page is about 22. Since most of the document span does not contain the answer, the number of negative samples (\textit{i.e.,} no answer) and positive samples (\textit{i.e.,} has answers) is extremely imbalanced. We follow~\cite{pan2019frustratingly,alberti2019bert} to sub-sample negative instances for training, where the rate of sub-sampling negative instance is the same as in~\cite{pan2019frustratingly}. As a result, there are 469,062 training instances in total.

We use Adam optimizer~\cite{kingma2014adam} with a batch size of $36$ for model training. The initial learning rate, the learning rate warmup proportion, the training epoch, the hidden size $h$, the number of blocks $T$, and the hyperparameter $K$ are set to $2 \times 10^{-5}$, $0.1$, $2$, $1024$, $2$, and $256$ respectively. Our model takes approximately 24 hours to train with 4 Nvidia Tesla P40. Evaluation completed in about 6 hours on the NQ dev and test set with a single Nvidia Tesla P100.

We use the Google released $\text{BERT-large}$ model fine-tuned with synthetic self-training~\cite{alberti2019synthetic} to encode the document and question as described in \S~\ref{sec:3.2.1}. We also compare the performance of RikiNet which uses the pre-trained RoBERTa\textsubscript{ large} model~\cite{liu2019roberta}.  
It should be noted that our RikiNet is orthogonal to the choice of a particular pre-trained language model.

\subsection{Main Results}
We present a comparison between previously published works on the NQ task and our RikiNet. We report the results of the precision (P), the recall (R), and the F1 score for the long-answer (LA) and short-answer (SA) tasks on both test set and dev set in Tab.~\ref{tab:main}. The first two lines of Tab.~\ref{tab:main} show the results of two multi-passage MRC baseline models presented in the original NQ paper~\cite{kwiatkowski2019natural}. The third to sixth lines show the results of the previous state-of-the-art models. These models all employ the BERT\textsubscript{large} model and perform better than that two baselines. Our RikiNet-BERT\textsubscript{large} also employs the BERT\textsubscript{large} model, and its single model has achieved a significant improvement over the previously published best model on the test set (LA from 66.8 F1 to 74.3 F1, and SA from 53.9 F1 to 57.9 F1). To the best of our knowledge, this is the first\footnote{The single RikiNet-BERT\textsubscript{large} model was submitted to the NQ public leaderboard on 7 Nov. 2019.} \textit{single} model that surpasses the single human performance~\cite{kwiatkowski2019natural} on both LA and SA tasks.
We also provide a BERT\textsubscript{joint}~\cite{alberti2019bert} + RoBERTa\textsubscript{ large}~\cite{liu2019roberta} baseline on NQ, which only replaces the BERT\textsubscript{large} in BERT\textsubscript{joint} method with RoBERTa\textsubscript{ large}. To be expected, the BERT\textsubscript{joint} + RoBERTa\textsubscript{ large} performs better than original BERT\textsubscript{joint}.
Furthermore, our single model of RikiNet-RoBERTa\textsubscript{ large} which employs RoBERTa\textsubscript{ large} model also achieves better performance on both LA and SA, significantly outperforming BERT\textsubscript{joint} + RoBERTa\textsubscript{ large}. These results demonstrate the effectiveness of our RikiNet.

Since most submissions on the NQ leaderboard are ensemble models, we also report the results of our ensemble model, which consists of three RikiNet-RoBERTa\textsubscript{ large} models with different hyper-parameters. At the time of submission (29 Nov. 2019), the NQ leaderboard shows that our ensemble model achieves the best performance on both LA (F1 76.1) and SA (F1 61.3).

\subsection{Ablation Study}
RikiNet consists of two key parts: DPDA reader and multi-level cascaded answer predictor. To get a better insight into RikiNet, we conduct an in-depth ablation study on probing these two modules. We report the LA and SA F1 scores on the dev set.   
\paragraph{Ablations of DPDA Reader} We keep the predictor and remove the component of the DPDA reader. The results are shown in Tab.~\ref{tab:dpda}. In (a), we remove the entire DPDA reader as introduced in \S~\ref{sec:dpcr} except BERT\textsubscript{large}. In (b), (c), and (d), we remove the dual-attention layer, question self-attention layer, and paragraph dynamic self-attention layer as described in \S~\ref{sec:3.2.1} respectively. In (e) and (f), we remove the paragraph attention mask of Eq. (\ref{eq:lmask}) and the dynamic attention mask of Eq. (\ref{eq:dmask}) respectively. 
We can see that after removing the DPDA reader, the performance drops sharply. In addition, the paragraph dynamic self-attention layer has the greatest impact on performance. Moreover, both the paragraph attention mask and dynamic attention mask contribute to the performance improvement.

We also change the hyper-parameter $K$ and the number of blocks $T$. Results show that the setting of $K=384$ performs better than $K=512$ (\textit{i.e.}, no dynamic attention mask), and $K=256$ performs best. For the number of DPDA blocks $T$, the model achieves the best performance when $T=2$.

\begin{table}[h]
\begin{center}
\small
  \begin{tabular}{lcccccl}
    \toprule
    Setting & LA F1 & SA F1 \\
    \midrule
    RikiNet-BERT\textsubscript{large} (Full) & \textbf{73.9} & \textbf{57.7}\\ \midrule
    (a) - DPDA reader & 70.7 & 55.9 \\
    (b) - Dual-attention layer & 73.1 & 56.6 \\ 
    (c) - Question self-attention layer & 73.5 & 57.5  \\ 
    (d) - Paragraph self-attention layer & 72.2 &
56.3  \\ 
    (e) - Paragraph attention mask & 73.2 &
57.1 & \\ 
    (f) - Dynamic attention mask & 72.9 &
56.8 \\ 
 \midrule
 RikiNet-BERT\textsubscript{large} ($K=512$) & 72.9 & 56.8 \\
 RikiNet-BERT\textsubscript{large} ($K=384$) & 73.7 &
57.3 \\
 RikiNet-BERT\textsubscript{large} ($K=256$) & \textbf{73.9} & \textbf{57.7} \\
 RikiNet-BERT\textsubscript{large} ($K=128$) & 73.7 & 56.9 \\
  \midrule
 RikiNet-BERT\textsubscript{large} ($T=0$) & 70.7 & 55.9 \\
 RikiNet-BERT\textsubscript{large} ($T=1$) & 73.6 & 57.6 \\
 RikiNet-BERT\textsubscript{large} ($T=2$) & \textbf{73.9} & \textbf{57.7} \\
 RikiNet-BERT\textsubscript{large} ($T=3$) & 73.5 &
57.1 \\
 RikiNet-BERT\textsubscript{large} ($T=4$) & 73.0 & 
56.9 \\
  \bottomrule
 
\end{tabular}
\end{center}
\caption{Ablations of DPDA reader on dev set of NQ dataset.}
\label{tab:dpda}
\end{table}

\paragraph{Ablations of Predictor} On the predictor side, we further remove or replace its component and report the results in Tab.~\ref{tab:predictor}. In (1) we remove the whole DPDA reader and predictor. In (2), we remove the way of multi-level prediction (\textit{i.e.}, training the model to predict long and short answer jointly) described in \S~\ref{sec:mcap}, and follow the previous work~\cite{alberti2019bert} to directly predict the short answer and then select its paragraph as the long answer. We can see that our multi-level prediction is critical to the long answer prediction. In (3) we only remove the cascaded structure but keep the multi-level prediction, which means that the prediction representations are no longer used as input for other predictions, the performance of both long and short answers drops about 1.0 F1 score. 
In (4) we change the ordering of cascaded process. That is instead of considering long answer first and then short answer as described in \S~\ref{sec:mcap}, we consider the cascaded structure of short answer first and then long answer. However, we get slightly worse results in this way. 
In (5), we remove the question embedding which is used for answer type prediction. It can be observed that the question embedding contributes to performance improvement.
In the variants of (6)-(9), we remove the dense prediction layers with $\Tanh$ activation function and replace it with Bi-directional Long-Short Term Memory (Bi-LSTM)~\cite{hochreiter1997long,schuster1997bidirectional} layers, transformer self-attention blocks, and dense prediction layers with Gaussian Error Linear Unit $\GELU$~\cite{hendrycks2016gaussian} activation function but neither get better performance.

Overall, both proposed DPDA reader and multi-level cascaded answer predictor significantly improve the model performance.

\begin{table}[t]
\begin{center}
\small
  \begin{tabular}{lcccccl}
    \toprule
    Setting & LA F1 & SA F1 \\
    \midrule
    RikiNet-BERT\textsubscript{large} (Full) & \textbf{73.9} & \textbf{57.7}\\ 
    \midrule
    (1) - DPDA reader \& Predictor & 65.9 & 55.1 \\
    (2) - Multi-level prediction & 70.9 & 57.1\\
    (3) - Cascaded structure & 73.0 & 56.7 \\
    (4) \ \ \ \ + S2L cascaded structure & 73.6 & 57.5 \\
    (5) - Question embedding & 73.4 & 57.4 \\
    (6) - Tanh dense prediction layer & 73.2 & 57.3 \\
    (7) \ \ \ \ + Bi-LSTM prediction layer & 73.3 & 57.4  \\
    (8) \ \ \ \ + Transformer prediction layer & 73.5 & 57.5 \\    
    (9) \ \ \ \ + GELU dense prediction layer & 73.7 & 57.6 \\
  \bottomrule
\end{tabular}
\end{center}
\caption{Ablations of multi-level cascaded predictor on dev set of NQ dataset.}
\label{tab:predictor}
\end{table}

\section{Related Works}

Natural Questions (NQ) dataset~\cite{kwiatkowski2019natural} has been recently proposed, where each question is paired with an entire Wikipedia page which is a long document containing multiple passages. 
Although BERT~\cite{devlin2018bert} based MRC models have surpassed human performance on several MRC benchmark datasets~\cite{lan2019albert,devlin2018bert,liu2019roberta,rajpurkar2018know}, a similar BERT method~\cite{alberti2019bert} still has a big gap with human performance on NQ dataset.

There are several recently proposed deep learning approaches for multi-passage reading comprehension. \citet{chen2017reading} propose DrQA which contains a document retriever and a document reader (DocReader). \citet{clark2018simple} introduce Document-QA which utilizes TF-IDF for paragraph selection and uses a shared normalization training objective. \citet{de2018question} employ graph convolutional networks (GCNs) for this task.
\citet{zhuang2019token} design a gated token-level selection mechanism with a local convolution. 
In contrast, our RikiNet considers multi-level representations with a set of complementary attention mechanisms.

To solve the NQ task, \citet{kwiatkowski2019natural} adapt Document-QA~\cite{clark2018simple} for NQ, and also utilizes DecAtt~\cite{parikh2016decomposable} for paragraph selection and DocReader~\cite{chen2017reading} for answer prediction. 
BERT\textsubscript{joint}\cite{alberti2019bert} modifies BERT for NQ. 
Besides, some works focus on using data augmentation to improve the MRC models on NQ. \citet{alberti2019synthetic} propose a synthetic QA corpora generation method based on roundtrip consistency. \citet{glass2019span} propose a span selection method for BERT pre-training (SSPT). 
More recently, \citet{pan2019frustratingly} introduce attention-over-attention~\cite{cui2016attention} into the BERT model. \citet{pan2019frustratingly} also propose several techniques of data augmentation and model ensemble to further improve the model performance on NQ.
Although the use of data augmentation and other advanced pre-trained language models~\cite{lan2019albert} may further improve model performance, as this is not the main focus of this paper, we leave them as our future work.
Our RikiNet is a new MRC model designed tailored to the NQ challenges and can effectively represent the document and question at multi-levels to jointly predict the answers, which significantly outperforms the above methods.

\section{Conclusion}
We propose the RikiNet, which reads the Wikipedia pages to answer the natural question. The RikiNet consists of a dynamic paragraph dual-attention reader which learns the token-level, paragraph-level and question representations, and a multi-level cascaded answer predictor which jointly predicts the long and short answers in a cascade manner. On the Natural Questions dataset, the RikiNet is the first single model that outperforms the single human performance. Furthermore, the RikiNet ensemble achieves the new state-of-the-art results at 76.1 F1 on long-answer and 61.3 F1 on short-answer tasks, which significantly outperforms all the other models on both criteria.

\section*{Acknowledgment}
This work is supported by National Natural Science Fund for Distinguished Young Scholar (Grant No. 61625204) and partially supported by the Key Program of National Science Foundation of China (Grant No. 61836006). 

\bibliographystyle{acl_natbib}
\bibliography{nq}

\end{document}